\documentclass[preprint,12pt]{elsarticle}
\usepackage[usenames,dvipsnames]{color}
\usepackage{amssymb}
\usepackage{amsmath}
 
\usepackage{makecell}
\usepackage{multirow}
\usepackage{array}
\usepackage{url}
\usepackage{caption}
\usepackage{subcaption}
\usepackage{colortbl}
\usepackage{lineno}
\usepackage{adjustbox}
\journal{Neurocomputing}

\begin{document}
%\noindent\hrulefill
\begin{frontmatter}



\title{EPAM-Net: An Efficient Pose-driven Attention-guided Multimodal Network for Video Action Recognition}


\author{Ahmed Abdelkawy, Asem Ali, and Aly Farag}
\ead{\{a0nady01, asem.ali, aly.farag\}@louisville.edu}
% \institute{Computer Vision and Image Processing Laboratory (CVIP), University of Louisville, Louisville, KY. \\
%\email{\{a0nady01, asem.ali, aly.farag\}@louisville.edu}}
\affiliation{organization={Computer Vision and Image Processing Laboratory (CVIP), University of Louisville},%Department and Organization
            %addressline={}, 
            city={Louisville},
            postcode={}, 
            state={KY},
            country={USA}}

%% Abstract
\begin{abstract}
Existing multimodal-based human action recognition approaches are computationally intensive, limiting their deployment in real-time applications. In this work, we present a novel and efficient pose-driven attention-guided multimodal network (EPAM-Net) for action recognition in videos. Specifically, we propose eXpand temporal Shift (X-ShiftNet) convolutional architectures for RGB and pose streams to capture spatio-temporal features from RGB videos and their skeleton sequences. The X-ShiftNet tackles the high computational cost of the 3D CNNs by integrating the Temporal Shift Module (TSM) into an efficient 2D CNN, enabling efficient spatiotemporal learning. Then skeleton features are utilized to guide the visual network stream, focusing on keyframes and their salient spatial regions using the proposed spatial-temporal attention block. Finally, the predictions of the two streams are fused for final classification. The experimental results show that our method, with a significant reduction in floating-point operations (FLOPs), outperforms and competes with the state-of-the-art methods on NTU RGB-D 60, NTU RGB-D 120, PKU-MMD, and Toyota SmartHome datasets. The proposed EPAM-Net provides up to a 72.8x reduction in FLOPs and up to a 48.6x reduction in the number of network parameters. The code will be available at \url{https://github.com/ahmed-nady/Multimodal-Action-Recognition}.
\end{abstract}

% \includegraphics[width=2\linewidth]{images/proposed_framework_journal.png}

% \begin{highlights}
% \vfill
% \item A novel and efficient pose-driven attention-guided multimodal network (EPAM-Net) is proposed for action recognition in videos.
% \item A pose-driven spatiotemporal attention block is introduced to guide the visual network stream focusing on discriminative frames and their salient human body regions. 
% \item Upto 60.8x reduction in FLOPs and 9.2x reduction in the number of network parameters compared to state-of-the-art architectures.
% \end{highlights}

%% Keywords

\end{frontmatter}


%\linenumbers
\section{Introduction}
\label{sec:intro}

 Human action recognition (HAR), which assigns an action class label to an input video segment, has been an active research area in computer vision. It plays a key role in several real-world applications such as video indexing and retrieval, visual surveillance, healthcare, patient rehabilitation, sports analysis, and measuring the behavioral engagement of students in classrooms \cite{karim2024human,karim2024hade,abdelkawy2024measuring,khalid2020supporting,garba2024understanding}. Recently, unimodal methods of HAR, such as skeleton-based or RGB video-based methods, have witnessed remarkable improvements.

The RGB video-based methods model the spatial-temporal representation from video data and its corresponding estimated optical flow using network architectures such as two-stream network \cite{karpathy2014large}, CNN-LSTM network \cite{donahue2015long}, and 3D convolutional neural networks (3D CNN) \cite{tran2015learning,qiu2017learning,carreira2017quo,feichtenhofer2019slowfast,feichtenhofer2020x3d}. Among these architectures, X3D is an efficient 3D CNN architecture, achieving competitive performance for action recognition. In this network architecture, a tiny 2D mobile CNN architecture, which utilizes channel-wise separable convolutions instead of standard ones, is progressively expanded along several axes: time, space, depth, and network width. Despite the efficiency of the X3D network, it still has higher computational demands compared to its 2D CNN counterpart.

On the other hand, the skeleton-based methods represent human action through the trajectories of body keypoints \cite{sun2022human}. Skeleton data can be obtained either from RGB videos using pose estimation algorithms or from motion capture systems, e.g., Kinect.  %( modeling its motion- the body posture configurations and their dynamics throughout time. 
 Skeleton-based approaches can be grouped into four categories based on the used network architecture: Convolutional Neural Network (2D-CNN) \cite{choutas2018potion,caetano2019skelemotion}, Recurrent Neural Network (RNN), Graph Convolutional Network (GCN) \cite{yan2018spatial,shi2019two}, and 3D CNN \cite{Duan_poseconv3d}. In 2D CNN-based methods \cite{choutas2018potion,caetano2019skelemotion}, manually designed transformations are utilized to model the skeleton sequence as a pseudo image, while RNN-based methods model temporal context within the skeleton sequence. Such input representations limit the exploitation of structural information of the skeleton sequence. GCN-based approaches \cite{yan2018spatial,liu2020disentangling} represent pose sequences as spatiotemporal graphs. However, the limitations of these approaches are their non-robustness to noises in pose estimation and the necessity of careful design for integrating the skeleton with other modalities. In contrast, in the 3D-CNN based method \cite{Duan_poseconv3d}, the input is represented as a volume of heatmaps, which capture the structure of skeleton joints and their dynamics over time. In this paper, we adopt the representation of the human skeleton sequence using a 3D pseudo-heatmap volume, and then the proposed X-ShiftNet is utilized to learn spatial-temporal representation from such pseudo heatmap volume.
  
\begin{figure*}[!t]
\includegraphics[width=\linewidth]{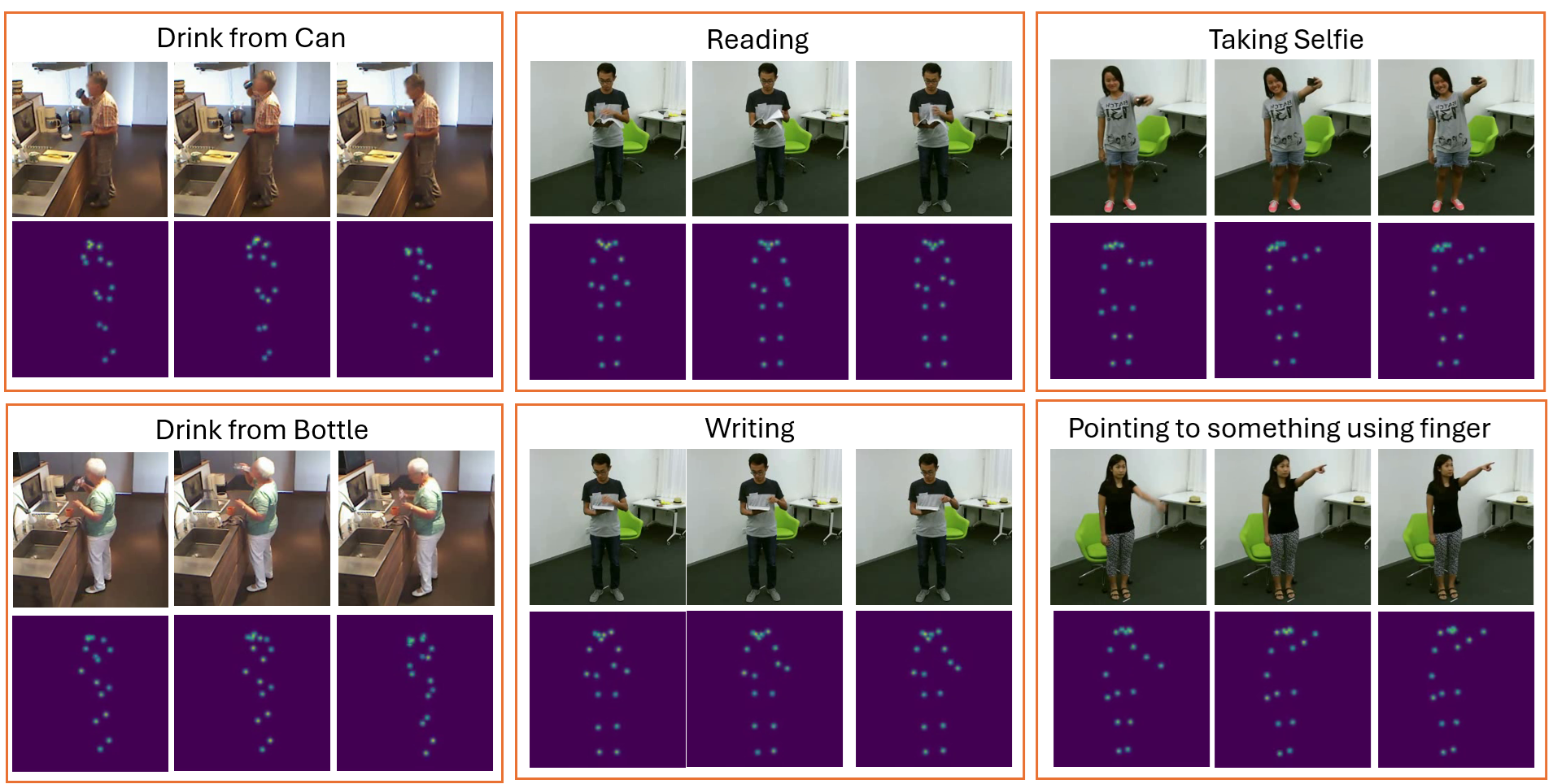}
\caption{An example of action pairs that are challenging to differentiate using the skeleton modality in Toyota-Smarthome (left) and in NTU-RGB+D dataset (middle and right), where the skeleton heatmaps of each pair of actions (e.g., reading-writing) are similar.}% Different actions may have similar heatmaps.}
\label{fig:motivation}
\end{figure*}

 RGB video and skeleton modalities offer distinct perspectives on human actions. The RGB modality provides detailed appearance information, including scene context and object interactions. However, RGB video-based approaches are vulnerable to changes in viewpoint, background, and illumination conditions. In contrast, human skeleton data represents actions as a sequence of moving skeleton joints, making it robust against the challenges of RGB-based approaches. Nevertheless, certain actions appear ambiguous when viewed solely through skeleton data because of lacking appearance details, such as interacting objects. Figure. \ref{fig:motivation} shows an example of action pairs that have similar skeleton movements, such as drinking from a can and drinking from a bottle, reading and writing, as well as pointing to something with a finger and taking a selfie. 

Therefore, multimodal HAR methods, which take advantage of the complementarity of RGB and skeleton modalities to improve the performance of HAR, have recently gained  attention \cite{baltruvsaitis2018multimodal}. Previous fusion-based multimodal HAR methods can be grouped into three categories: score-level, feature-level, and model-level fusion. Score fusion-based models handle the skeleton and RGB information separately and then aggregate their scores from Softmax layers. However, RGB and skeleton modalities fail to boost each other for feature representation. On the other hand, feature fusion-based approaches concatenate modalities' features at the fully connected layers of modality-specific models. These methods slightly improved the HAR performance due to not considering the alignment of RGB data and its corresponding human body poses. Das et al. \cite{das2020vpn} addressed such an alignment problem through proposing spatial embedding, which projects visual features and 3D skeletons in the same referential. Moreover, the temporal alignment is performed by assuming the existence of a 3D pose for each frame. However, this approach is computationally expensive. In contrast, model-level fusion-based methods employ the knowledge from one data modality to facilitate modeling in other data modalities \cite{bruce2022mmnet}.  Bruce et al. \cite{bruce2022mmnet} used skeleton modality to learn spatial attention and then weight the spatiotemporal region of interest (ST-ROI) map, which is constructed from video input, accordingly. Although this approach uses a 2D CNN to learn visual features from the ST-ROI map, it is still computationally intensive.

To tackle the mentioned limitations, we propose a novel Efficient Pose-driven Attention-guided Multimodal Network (EPAM-Net) that exploits the complementarity of RGB and skeleton modalities by aligning skeleton sequences and RGB videos in spatial and temporal dimensions. EPAM-Net consists of two eXpand temporal Shift networks (X-ShiftNets), which integrate the Temporal Shift Module (TSM) into an efficient 2D CNN, enabling efficient spatiotemporal learning, and a lightweight spatiotemporal attention block.
In summary, our main contributions are as follows:
 \begin{enumerate}
     \item We introduce two X-ShiftNet models, which achieve and surpass efficient 3D CNNs (e.g., X3D network) performance while requiring 1.1x and 1.4x fewer FLOPs and network parameters.
     \item We introduce a pose-driven spatiotemporal attention block to guide the visual stream focusing on discriminative frames and their salient human body regions.
    \item Using the aforementioned two components, we construct an efficient multimodal architecture (EPAM-Net) that exploits the complementarity of the RGB and skeleton modalities in an end-to-end manner. 
     \item Our EPAM-Net outperforms and competes with state-of-the-art methods on four benchmarks while reducing FLOPs and network parameters up to 72.8x and 48.6x, respectively.
 \end{enumerate}
 
\section{Related work}
In this section, we review the HAR literature according to the model's type: unimodal HAR (RGB video-based or skeleton-based HAR) and multimodal HAR.
 \subsection{RGB video-based action recognition approaches}
 Due to the easy collection of RGB data, RGB video-based methods have rapidly developed and obtained impressive results. RGB video-based methods can be divided into three categories \cite{sun2022human}: Two-stream 2D CNN-based, RNN-based, and 3D CNN-based methods. Two-stream 2D CNN methods \cite{simonyan2014two} comprise two 2D CNN streams to learn the appearance and motion features from RGB video and its corresponding estimated optical flow. RNN-based methods \cite{donahue2015long} extract frame-level visual features using 2D CNN, then utilize gated RNN architectures, e.g., Long-Short Term Memory (LSTM) to capture the long-term temporal dynamic in a video sequence. To concurrently learn the spatial and temporal information from RGB video, 3D CNN-based methods are introduced. Two-stream Inflated 3D CNN \cite{carreira2017quo} was introduced through temporally extending the convolutional and pooling kernels of a 2D CNN. Feichtenhofer et al.\cite{feichtenhofer2019slowfast} proposed a SlowFast network with two paths that operate on video frames but at two different speeds. The slow pathway, which works at a low frame rate, models spatial semantics, while the fast pathway models fine motion by working at a high frame rate. The lateral connection is used to fuse the two pathways. Despite the impressive results of 3D CNN-based methods, they require heavy computation to extract spatio-temporal features from videos. As a result, Temporal Shift Module (TSM) \cite{lin2019tsm} is proposed to enable 2D CNN networks to achieve 3D CNN performance without adding computational overhead. TSM facilitates temporal interactions among features of neighboring frames through shifting a part of the channels along the temporal dimension. Feichtenhofer \cite{feichtenhofer2020x3d} introduced an efficient 3D CNN architecture (X3D) for action recognition by progressively expanding a tiny 2D mobile image classification architecture into a spatiotemporal one along several possible axes: temporal duration $\gamma_{t}$, spatial resolution $\gamma_{s}$, network depth $\gamma_{d}$, network width $\gamma_{w}$, bottleneck width $\gamma_{d}$, and frame rate $\gamma_{\tau}$. This progressive expansion starts by expanding the bottleneck width of the model $\gamma_{d}$, followed by frame rate $\gamma_{\tau}$, spatial resolution $\gamma_{s}$, depth of the network $\gamma_{d}$, temporal/duration $\gamma_{t}$, and finally, global width of the layers $\gamma_{w}$. Due to the efficiency and competitive performance of the X3D network, we built upon it to develop our X-ShiftNet model.  
\subsection{Skeleton-based action recognition approaches}
Skeleton-based approaches can be grouped into four classes based on architectures: 2D CNN, RNN, GNN, and 3D CNN.  For CNN, Choutas et al.~\cite{choutas2018potion} introduced video clip-level human pose-based representation that encodes body joints’ motion during the entire clip. The pose motion representation (PoTion) is constructed by temporally aggregating the heatmaps of each joint by colorizing each of them based on their order in the video clip. After that, PoTion is used as input for the proposed 2D-CNN network architecture to predict the action category.
Liu et al.~\cite{liu2018recognizing} observed that using pose estimation maps that maintain more details of human body shape is more beneficial for action recognition than depending solely on the inaccurate 2D coordinates of body joints. So, they generated two evolution maps: a body shape evolution map from a sequence of averaged body joints heatmaps and a body joints evolution map from a sequence of pseudo-heatmaps of estimated body joints 2D coordinates. The two evolution maps are used as input for the proposed two-stream 2D CNN architecture, and then the prediction label is obtained by averaging the two CNN scores.

For GCN-based approaches, Yan et al.\cite{yan2018spatial} proposed the Spatial-Temporal Graph Convolutional Network (ST-GCN) to model the dynamics of skeleton sequence and the spatial arrangements of its joints. They created a spatial graph by utilizing the inherent connections between joints in the human body, and they also introduced temporal edges between corresponding joints in consecutive frames to extend the spatial graph into the spatiotemporal domain.  
The limitation of this work is that the skeleton GCN's graph is heuristically preset and it reflects only the human body's physical structure. It also remains fixed across all layers and input samples.  Lie et al.\cite{shi2019two} addressed this limitation by introducing an adaptive graph convolutional network that is capable of adaptively learning the graph topology for different GCN layers and skeleton samples. To improve the accuracy of classification, the joints and second-order information of the skeleton represented in the bone lengths and their directions were modeled using a two-stream adaptive graph convolutional network (2s-AGCN).

For 3D CNN approaches, Duan et al.~\cite{Duan_poseconv3d} represented the sequence of the human skeleton using a 3D pseudo-heatmap volume. Then, a 3D CNN network called slowOnly is used to classify the 3D heatmap volumes into one of the action categories. Compared to the SlowOnly model \cite{Duan_poseconv3d}, we develop X-ShiftNet model to learn spatiotemporal information from the pseudo-heatmap volume of the skeleton sequence.

\subsection{Multimodal action recognition approaches}
Multimodal HAR approaches can be grouped into two categories: fusion and distillation-based approaches. Fusion-based approaches can be further classified into three groups: score-level fusion, feature-level fusion, and model-level fusion. 
In score fusion, the skeleton and RGB information are modeled separately, and then the classification scores of both network streams are fused to obtain the final prediction. On the other hand, feature fusion-based methods concatenate modality-specific features either at the fully connected layers of modality-specific models or at several layers using lateral connections \cite{Duan_poseconv3d}. Zolfaghari et al.~\cite{zolfaghari2017chained} introduced a deep network architecture to utilize the three visual cues (pose, motion, and appearance) and fuse them sequentially using a Markov chain model. Each modality has its 3D-CNN architecture (C3D). In the chained architecture, the prediction of each stream relies not only on the stream input but also on all predictions of previous streams. As a result, the network in that stream refines class labels from prior streams by learning complementing features. 
 Li et al. \cite{li2020sgm} introduced a skeleton-guided multimodal network (SGM-Net) to exploit the complementarity of RGB and skeleton modalities at the feature level. The network is composed of three components: ST-GCN \cite{yan2018spatial} to extract pose features, the R(2+1)D network \cite{tran2018closer} to extract visual features, and a guided block to pay attention to action-related features in RGB videos. Unlike SGM-Net, where visual features are extracted from the whole spatial resolution of video frames, we extract these features from cropped human regions to reduce the interference of background in action classification. Das et al. \cite{das2020vpn} introduced video-pose network (VPN), whose input is 64 video frames and their corresponding 3D poses. They used the I3D network to extract spatiotemporal features from these 64 frames, while the GCN model was employed to learn pose features from 3D poses. The pose features are used to learn spatial-temporal attention, and then spatial embedding is proposed to find the correspondence between human joints and their relevant image regions to modulate visual features, which are used for classification. Unlike VPN \cite{das2020vpn}, our proposed approach represents a skeleton sequence using a pseudo heatmap volume, which is spatially aligned with a cropped RGB human region. %generated according to the shared minimum bounding box involving all 2D poses across video.  

For model-based fusion, Bruce et al. \cite{bruce2022mmnet} proposed a model-based multimodal network (MMNet) that learns spatial attention from the skeleton modality using a GCN network and then weights the spatiotemporal region of interest (ST-ROI) map accordingly. An ST-ROI map is constructed by cropping the frame's body area of the actor(s) head, hands, and feet from five sampled frames of a video input. After that, such ST-ROI is weighted and fed to 2D CNN to learn the appearance feature. Unlike MMNet \cite{bruce2022mmnet}, which uses spatial attention on ST-ROI, not spatial-temporal attention that may not assign different weights for each body part through frames, we use person-centered modeling and spatial-temporal attention to modulate visual features accordingly. 

Conversely, distillation-based approaches incorporate pose information into a network with RGB input, eliminating the need for poses at inference. Das el at. \cite{das2021vpn++} proposed VPN++ network to augment the RGB representation with 3D pose information through features and attention-level distillation. However, it neglected the alignment between 2D skeleton joints and their corresponding spatial regions in RGB frames. To address the alignment problem, Reilly and Das \cite{reilly2024just} proposed 2D and 3D pose induction modules to integrate 2D and 3D pose information into TimeSFormer with RGB input. However, their method is computation-intensive with 590.0 GFLOPs. 

In this work, we propose a multimodal action recognition approach that addresses these limitations and reduces the computation complexity as well as enhances the performance by developing X-ShiftNet to learn spatial-temporal features and modulating visual features using spatial-temporal attention. Table \ref{tab:person-centric_comparison} shows a comparison of our EPAM-Net with existing person-centric multimodal HAR approaches in terms of modality interactions, contributions, and limitations.

\begin{table}[]
 
\footnotesize
\caption{ Comparison between person-centric multimodal HAR approaches.}
    \label{tab:person-centric_comparison}
    \centering
    \begin{adjustbox}{width=.95\columnwidth,center}
    \begin{tabular}{p{1.7cm} p{2cm} p{6cm} p{5cm}}
    \hline
         Work  & Modality interactions &Contributions & Limitations  \\ \hline
         
 VPN\cite{das2020vpn}& Feature-level fusion & $\diamond$ Learn spatial and temporal attention weights separately, then fuse them via a Hadamard product.
        $\diamond$ Align 3D joints with image regions via spatial embedding. &$\diamond$ Intensive computation (107.9 GFLOPs). $\diamond$ Performance depends on 3D pose quality. \\ \hline

VPN++\cite{das2021vpn++} & Knowledge distillation &$\diamond$) Augment RGB representation with 3D pose information through features and attention-level distillation. & $\diamond$ Intensive computation (125.8 GFLOPs). $\diamond$ Ignoring skeleton-image alignment. $\diamond$ Comparable results demand 3D poses at inference. \\ \hline
MMNet\cite{bruce2022mmnet} & Model-level and score-level fusion & $\diamond$ Construct ST-ROI from five-RGB frame actors' head/hands/feet crops. \, $\diamond $ Utilize  GCN-guided spatial attention to weigh the ST-ROI map before 2D CNN classification. &$\diamond $ Use spatial, not spatiotemporal, attention on ST-ROI, thus lacking per-part, per-frame weighting. $\diamond $ Performance depends on 3D pose quality.  \enspace $\diamond $ High computational cost (89.2 GFLOPs). $\diamond $ Need 2D pose to crop head/hands/feet from RGB frames and 3D pose for GCN-based pose stream input.\\ \hline

TCEM-MMNet\cite{liu2024temporal}& Model-level and score fusion &$\diamond$ Introduce TCEM, composed of ResNet18 and a 3-layer LSTM for spatial and temporal feature learning of ST-ROI. & $\diamond$Intensive computation (85.4 GFLOPs).$\diamond$ Performance depends on 3D pose quality. $\diamond$ Need 2D pose to crop head/hands/feet from RGB frames and 3D pose for GCN-based pose stream input.
\\ \hline

$\pi$-ViT\cite{reilly2024just}&  Knowledge distillation &$\diamond$ Integrate 2D/3D pose information into TimeSFormer’s RGB backbone via 2D/3D pose induction module. & $\diamond$ High computation cost (590.0 GFLOPs). $\diamond$ Comparable results demand 3D poses at inference. \\ \hline

Ours & Feature-level and score-level fusion & $\diamond$ Introduce X-ShiftNets to learn spatiotemporal features from aligned RGB frames and pose sequences. $\diamond$ Introduce a lightweight nesting spatiotemporal attention block. $\diamond$ EPAM-Net rivals SOTA methods with 8.1 GFLOPs.
& $\diamond$ Performance depends
on 2D pose quality. \\ \hline

    \end{tabular}
    \end{adjustbox}
 
\end{table}
 
\section{The proposed EPAM-Net}
\begin{figure*}[!t]
\includegraphics[width=\linewidth]{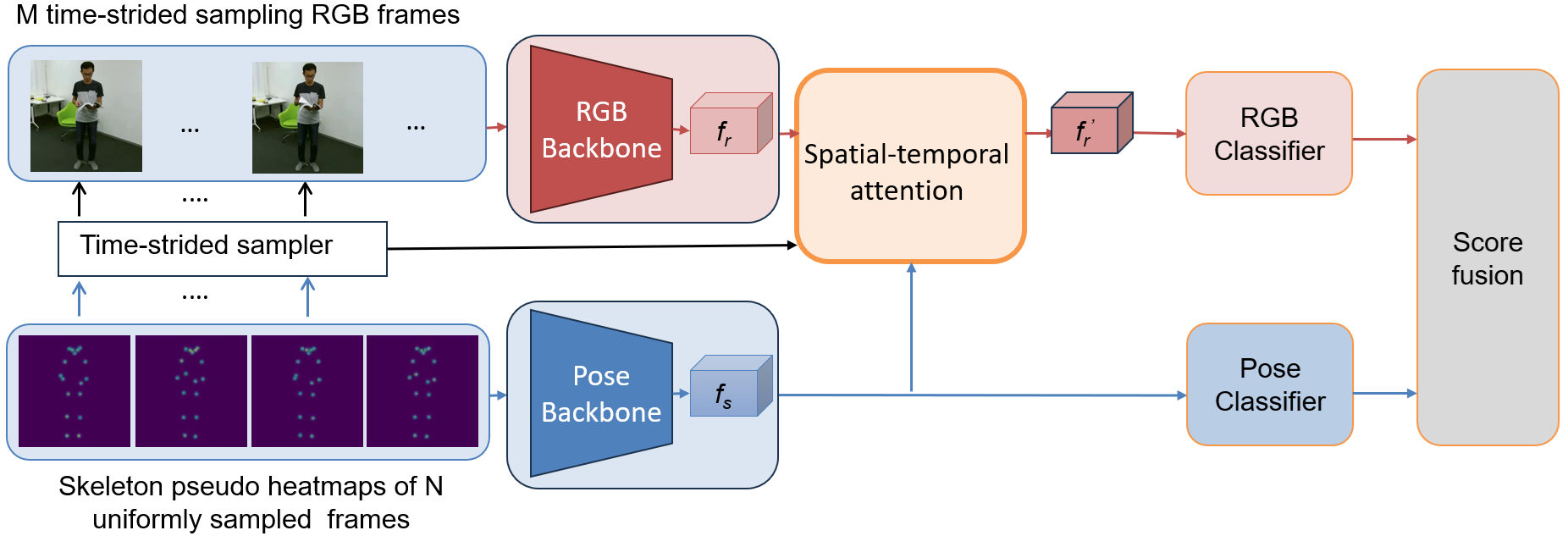}
\caption{The EPAM-Net architecture consists of visual and pose backbones to extract spatial-temporal features from RGB videos and pose sequences, respectively; a pose-driven spatial-temporal attention block to re-weight visual features accordingly; two classification heads; and final score fusion. The input of the pose network stream is a pseudo-heatmaps volume from N uniformly sampled frames, while the input of the visual network stream consists of M frames selected from these N frames by sampling one out of every $\frac{N}{M}$ frame. $f_s$ and $f_r$ represent skeleton features and visual features, respectively.} 
\label{fig:proposedFramework}
\end{figure*}

 Our proposed approach focuses on person-centric modeling by capturing both human body movements and interacting objects. To achieve this, we first determine the minimum bounding box that encompasses all 2D human skeletons across video frames. Then, each frame is cropped according to this bounding box and resized to the target dimensions. Figure. \ref{fig:proposedFramework} illustrates an overview of the proposed network architecture. The input to our proposed network is cropped RGB frames and their corresponding skeleton sequence. Specifically, the pose stream input is the pseudo-heatmap volume constructed from N uniformly sampled frames from the input video clip, while the RGB stream input consists of M frames selected from these N frames by picking one frame out of every $\frac{N}{M}$ frames.  
 
 The pipeline of the proposed multimodal network includes: 1) Extracting the spatiotemporal dynamics of skeleton sequences and cropped RGB frames separately using the proposed X-ShiftNet models; 2) Guiding the visual network stream to focus on discriminative human body part(s), interacting objects, and keyframes a using spatial-temporal attention block; 3) Fusing the class scores of the two streams of the proposed network for further performance improvement. The rationale behind this pipeline is that the M-frame RGB video input (e.g., M=16) might not be sufficient to capture the full temporal dynamics of an action. In contrast, the pose network, which utilizes N  uniformly sampled frames (e.g., N=48 > M), can capture more temporal information. Below we discuss the visual network, pose network, and spatial-temporal attention block in detail.
 
\subsection{Visual stream}
The X-ShiftNet network is proposed to extract spatiotemporal features $F_r \in \mathbb{R}^{C\mathrm{x}T\mathrm{x}H\mathrm{x}W}$ from RGB video frames, where C represents the number of channels, T the number of frames, and HxW the spatial resolution.
The X-ShiftNet network, inspired by TSM \cite{lin2019tsm} and X3D networks \cite{feichtenhofer2020x3d}, achieves the effect of 3D convolution using 2D convolution by moving a portion of the input feature channels along the temporal dimension. 
The proposed X-ShiftNet combines the architectural design strength of the X3D network and the temporal modeling capability of TSM, capturing spatio-temporal features without increasing network parameters or computation overhead. Specifically, we obtain the 2D CNN counterpart of the X3D network by removing the temporal convolution in the conv1 stage and replacing all 3D convolutions with 2D convolutions. Then, TSM is added for each inverted residual block of the network. Moreover, we employ one fully connected layer instead of two for the classification head. Table \ref{table:x3dTShift}. shows the instantiating of the RGB-X-ShiftNet network.
  
\subsection{Pose stream}
\label{sec:pose_stream}
We develop an X-ShiftNet-s network to extract spatiotemporal dynamics from skeleton sequences. The proposed RGB-X-ShiftNet network is modified for skeleton-based action recognition as follows: 1) Remove the first stage, and 2) Change the spatial stride of the first convolution layer of the stem layer from 2 to 1. This makes the spatial resolution of the final feature map match that of the visual backbone feature maps. Table \ref{table:x3dTShift}. shows the instantiation of the Pose-X-ShiftNet network. 

 Following the work \cite{Duan_poseconv3d}, we employ the Top-Down pose estimation approach \cite{Duan_poseconv3d} with Faster-RCNN 
 as a detector and HRNet \cite{sun2019deep} as a pose estimation. Having skeleton coordinate triplets ($x_k,y_k,c_k)$ for each frame, the K heatmaps are generated using a Gaussian map centered at each joint:
\begin{equation} H_k(x,y) = \exp{- \frac{(x-x_k)^2 + (y-y_k)^2}{2\sigma^2}}\end{equation}
where $x_k$ and $y_k$ are the coordinates of $k$th joint and $\sigma$ controls Gaussian map variance.
Finally, all K joints heatmaps are stacked along the temporal dimension to form the 3D heatmap volume with size $K\mathrm{x}T\mathrm{x}H\mathrm{x}W$ where K is a number of human body keypoints, T is temporal length, and H and W are the height and width of such maps.  
\delimiterfactor=500
\delimitershortfall=10pt
\begin{table}
\caption{X-ShiftNet architectures for RGB and Pose streams. The kernel dimensions are represented by $T \mbox{x} S^2, C$ for temporal, spatial, and channel sizes, respectively. TSM is the temporal shift module.}% GAP stands for global average pooling.} %ME stands for motion excitation module.}
\label{table:x3dTShift}
\centering
\footnotesize
\begin{tabular}{ c|c|c|c } 
 \hline
 stage & RGBNet&PoseNet & output sizes $T \mbox{x} H \mbox{x} W$\\ \hline
 data layer & & &\makecell{RGB: 16x224x224\\
                           Pose: 48x56x56} \\ \hline
 conv1 & $1 \mbox{x} 3^2, 24$  & $1 \mbox{x} 3^2, 24$&\makecell{ RGB: 16x112x112 \\
                                                                 Pose: 48x56x56}\\ \hline
 res2 &  $   \left[  \makecell{TSM\\
            1 \mbox{x} 1^2 , 54 \\
            1 \mbox{x} 3^2, 54 \\
            1 \mbox{x} 1^2, 24 }\right]   x3 $ & 
            None 
            & \makecell{ RGB: 16x56x56 \\ 
                            Pose: 48x56x56 }\\ \hline %\fbox{ME}
res3 & $ \left[  \makecell{TSM\\
            1 \mbox{x} 1^2 , 108  \\
             1 \mbox{x} 3^2, 108  \\
             1 \mbox{x} 1^2, 48\\ 
               }\right]  x5$ & 
                $ \left[ \makecell{ TSM\\
            1 \mbox{x} 1^2 , 54 \\
            1 \mbox{x} 3^2, 54 \\
            1 \mbox{x} 1^2, 24   }  \right]   x5$
                &\makecell{ RGB: 16x28x28 \\ 
                            Pose: 48x28x28 }\\ \hline
res4 & $\left[  \makecell{TSM\\
            1 \mbox{x} 1^2 , 216  \\
             1 \mbox{x} 3^2, 216  \\
             1 \mbox{x} 1^2, 96\\
             }\right]  x11 $ &
             $ \left[ \makecell{  TSM\\
            1 \mbox{x} 1^2 , 108  \\
             1 \mbox{x} 3^2, 108  \\
             1 \mbox{x} 1^2, 48} \right]  x11$
             &\makecell{ RGB: 16x14x14 \\
                         Pose: 48x14x14 }\\ \hline
res5 & $\left[  \makecell{TSM\\
                1 \mbox{x} 1^2 , 432  \\
             1 \mbox{x} 3^2, 432  \\
             1 \mbox{x} 1^2, 192 \\
              }\right] x7$  &
              $\left[ \makecell{TSM\\
                1 \mbox{x} 1^2 , 216  \\
             1 \mbox{x} 3^2, 216  \\
             1 \mbox{x} 1^2, 96} \right] x7$
              &\makecell{ RGB: 16x7x7\\
                          Pose: 48x7x7} \\ \hline
conv5 &  $1 \mbox{x} 1^2, 432 $&$ 1 \mbox{x} 1^2, 216$& \makecell{ RGB: 16x7x7\\
                                                                        Pose: 48x7x7} \\ \hline

\multicolumn{3}{c|} {  
            global average pooling, fc} & \#classes\\ \hline

 \end{tabular}
 
\end{table}
 
\subsection{Spatial-temporal attention block}
\begin{figure}
    \centering
    \includegraphics[width=1\linewidth]{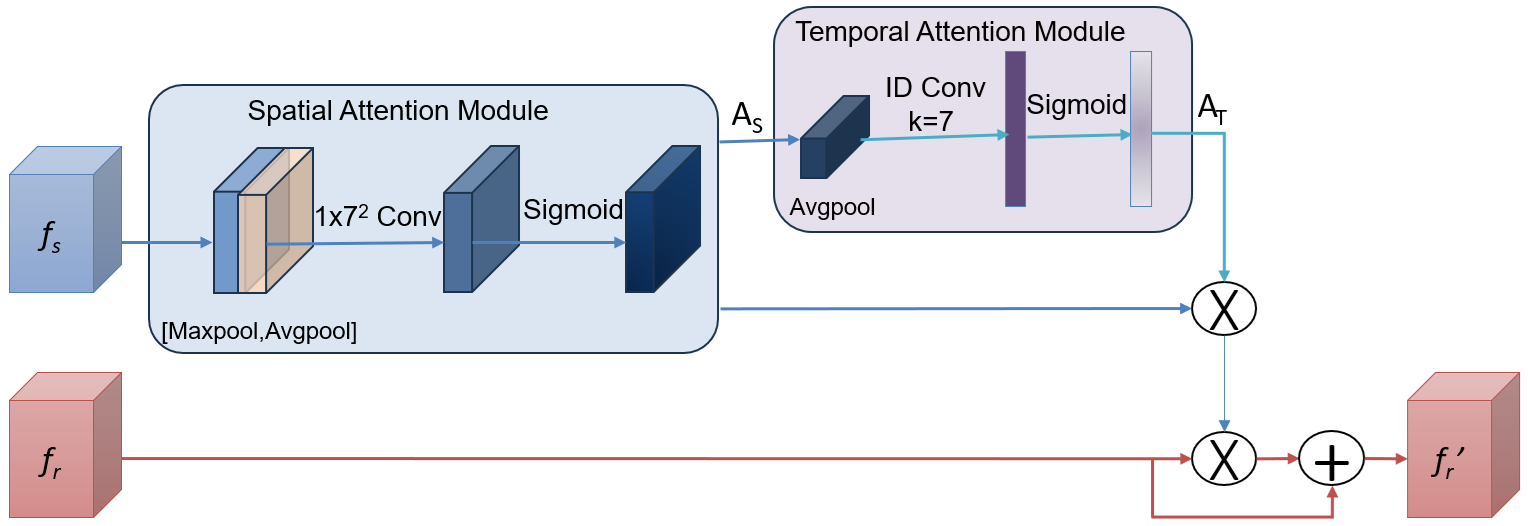}
    \caption{Illustration of the proposed spatial-temporal attention block. A spatial attention map weights discriminative spatial regions, while a temporal attention map weights keyframes.} 
    \label{fig:attention_block}
\end{figure}

After extracting skeleton features using the Pose-X-ShiftNet network and visual features using the RGB-X-ShiftNet network, the nesting spatial-temporal attention block, as shown in Fig. \ref{fig:attention_block}, is proposed to learn which spatial regions in each frame and which frames are worth paying attention to using skeleton features and then weigh visual features accordingly. The nesting spatial-temporal attention block consists of a spatial attention module, followed by a nested temporal attention module, which uses the spatial attention map as an input. Our nesting spatiotemporal attention, like VPN's spatiotemporal coupler \cite{das2020vpn}, modulates visual features by assigning different weights to each frame and its spatial regions. However, rather than computing spatial and temporal attentions separately as in VPN \cite{das2020vpn}, we explicitly model their interaction, gaining an advantage of attention to attention. This contrasts with MMNet \cite{bruce2022mmnet} as well, which uses skeleton joint weights (i.e., spatial attention) to weight the ST-ROI map, the input of the 2D CNN of the RGB stream.

To properly use spatial-temporal attention, we align skeleton pseudo-heatmaps with corresponding RGB frames. %For spatial alignment, we do the same preprocessing step to both RGB and pose input. 
 For spatial alignment, video frames are cropped according to the minimum bounding box involving all 2D human skeletons across the video frames, and then the skeleton pseudo-heatmap volume is generated accordingly. Moreover, the spatial resolution of the final feature maps for visual and pose backbones is matched to ensure spatial correspondence between the two modalities. For temporal alignment, since the RGB and pose modalities have different temporal resolutions, their feature maps should be matched in time for accurate action recognition. In particular, denoting the shape of the pose feature as $\{C_p,T_N,S^2\}$, and the shape of the RGB feature as $\{C_r,T_M,S^2\}$, so, $T_N$ is aligned with $T_M$ through time-strided sampling. Below, we discuss the spatial and temporal attention module in detail.
 
\subsubsection{Spatial attention module}
Given the skeleton feature maps $F_s \in \mathbb{R}^{C\mathrm{x}T\mathrm{x}H\mathrm{x}W}$,  
the spatial attention map $A_S \in \mathbb{R}^{1\mathrm{x}T\mathrm{x}H\mathrm{x}W}$ is obtained by compressing channel-wise features using max-pooling and average-pooling operations, followed by a 1x7x7 convolution. Specifically, the process of spatial attention can be expressed as follows:
\begin{equation}
   A_S = \phi(g^{1\mathrm{x}7\mathrm{x}7}([GAP(F_s);GMP(F_s)]))  
\end{equation}
 where GAP denotes global average pooling, GMP represents the global max pooling.
 
The spatial attention map reveals the importance of each spatial region in each video frame, with those of larger weights representing discriminative regions for the action.
\subsubsection{Temporal attention module}
The temporal attention module is inspired from \cite{wang2020eca}. It has two operations: a squeeze operation, in which global average pooling is used to aggregate spatial dimensions of a spatial attention map $A_S$, and an excitation operation, in which a 1D convolution models temporal-wise interactions among neighboring frames. Overall, the two operations of the temporal attention block can be formulated as: 
\begin{equation} Z_t = \frac{1}{W \mbox{x} H} \sum_{i=1}^{W}\sum_{j=1}^{H} A_S(:,i,j,:) ,\end{equation} 
\begin{equation}
A_T = \phi(Conv1D(Z_t)\end{equation}
where Conv1D is 1D Conv with a kernel size of 7.  

 The temporal attention map $A_T$ represents the importance of the T frames, with frames having larger weights in $A_T$ expected to be keyframes. The spatiotemporal attention map $A_{ST}$ is obtained by multiplying spatial attention map $A_S$ and temporal attention map $A_T$, i.e, $A_{ST} = A_S\otimes A_T$. 
 After that, the RGB feature is modulated according to $A_{ST}$ as follows:
 $f_{r}^{'} = f_r + f_r\otimes A_{ST}$. The reason for adopting a residual connection in modulating the RGB feature is the low quality of 2D pose estimation due to occlusion, low resolution, and truncation for some datasets, e.g., Toyota-Smarthome dataset.   
 
\section{Experiments}
We evaluate the proposed multimodal network on NTU RGB+D 60 \cite{shahroudy2016ntu}, NTU RGB+D 120  \cite{liu2019ntu}, PKU-MMD \cite{liu2017pku}, and Toyota-Smarthome \cite{das2019toyota} datasets. We report the mean Top-1 accuracy for Toyota-Smarthome dataset \cite{das2019toyota} following \cite{bruce2022mmnet,reilly2024just} and Top-1 accuracy for other datasets using 1-clip per video.

\subsection{Datasets}
 \textbf{NTU RGB-D dataset \cite{shahroudy2016ntu,liu2019ntu}}. The dataset is a large-scale, multi-modalities human action recognition dataset captured in a lab-controlled environment. It is available in two variants, NTU-60 and NTU-120. 
 The NTU RGB-D 60 dataset has 56,880 video clips of 60 human actions performed by 40 volunteers, whereas the NTU RGB-D 120 dataset has 114,480 videos of 120 human actions performed by 106 volunteers. Each action is simultaneously captured from three distinct horizontal views for several camera setups.  Each camera setup has a different height, and the three cameras are positioned at that height. The datasets have three settings for evaluation: cross-subject (X-Sub), cross-view (X-View for NTU 60), and cross-setup (X-Set for NTU-120).  In cross-subject (X-Sub), half of the subjects are used for training and the other half for testing. For X-View, video samples are split based on camera IDs (cameras 2 and 3 for training, camera 1 for testing), while X-Set splits are based on camera setups (even setup IDs for training and odd ones for testing).
\newline
\textbf{Toyota-Smarthome dataset \cite{das2019toyota}}. This dataset contains activities of daily living (ADL) collected in a smart home where 18 elderly people perform daily living tasks spontaneously. This dataset comprises 16,115 video clips with 31 activity classes captured from 7 viewpoints. The dataset evaluation protocols are cross-subject (CS) and cross-view (CV2) \cite{das2021vpn++}. The cross-view (CV1) evaluation setting is ignored due to the small number of training samples. 
\newline
\textbf{PKU-MMD dataset \cite{liu2017pku}}. The dataset comprises 21,545 action samples in 51 action categories performed by 66 participants and recorded simultaneously from left, center, and right viewpoints using three Microsoft Kinect v2 cameras. It employs two evaluation protocols: Cross-Subject (x-sub), where 57 subjects are used for training and 9 for testing, and Cross-View (x-view), where data from the center and right cameras are used for training, while the left camera is reserved for testing.
 
\subsection{Implementation details}

For skeleton modality, following the work \cite{Duan_poseconv3d}, we utilize a Top-Down pose estimation approach instantiated with HRNet-W32 \cite{sun2019deep} to extract 2D poses from videos and save the coordinate triplets (x, y, score). We then generate the pseudo heatmaps volume, the input of the pose stream, as follows: 1) 48 frames are sampled uniformly from the video. 2) 17 heatmaps (one heatmap per joint) are generated for each sampled frame (see Section. \ref{sec:pose_stream}) and then all such heatmaps are stacked along the temporal dimension. 3) The heatmaps are cropped with the global box that envelops all persons in the video and resized to 56 x 56. For RGB video modality, we select 16 frames from these 48 sampled frames of the pose stream via time-strided sampling. We then crop such frames with the global box and resize them to a resolution of 224 x 224. The cropped RGB frames are used as input for the RGB stream. Horizontal flipping is applied as data augmentation during training for RGB and pose streams.   

The training process of the proposed multimodal network involves two phases: first, pre-training RGB and skeleton networks, followed by fine-tuning the entire multimodal network for final classification. The RGB stream (RGB-X-ShiftNet) is trained with a batch size of 64 and a learning rate of 0.05 for 200 epochs, while the skeleton stream (Pose-X-ShiftNet) is trained with a batch size of 96 and a learning rate of 0.0375 for 240 epochs. The entire multimodal network is trained with a batch size of 24 and a learning rate of 0.001 for 10 epochs. The optimizer used to train all networks is the Stochastic Gradient Descent (SGD) with a momentum of 0.9. The learning rate is decreased using the cosine annealing schedule. The loss function of the proposed multimodal network is the summation of two cross-entropy losses of RGB and skeleton streams. For X-ShiftNets, we empirically shift 1/8 of feature channels forward and another 1/8 backward in the RGB stream, while in the Pose stream, we shift 1/4 forward and another 1/4 backward, as detailed in Section \ref{sec:portion_shift}. On Toyota-Smarthome dataset, we pretrain the two-stream X-ShiftNets on the NTU RGB-D 120 dataset since training sets on cross-subject and cross-view2 protocols are 8761 and 7685 video clips, respectively. We adopt PyTorch for implementation and train the RGB and Pose streams on 8 NVIDIA Tesla V100 PCIe 16 GB and finetune the proposed multimodal network on 2 NVIDIA TITAN RTX GPUs 24 GB.   

 \subsection{Ablation studies}
 In this section, we assess the effectiveness of each component of the proposed pose-driven attention multimodal architecture. Moreover, we compare the performance of network architectures for RGB and pose streams according to the complexity/accuracy trade-off. Finally, we evaluate the effectiveness of the nesting spatial-temporal attention block.
 \subsubsection{Effectiveness of the proposed multimodal architecture components}
 From Table \ref{table:ablation_study_proposed_method}, we can notice that RGB video-based network has higher top-1 accuracies compared to its skeleton-based counterpart on the NTU RGB-D 60 and NTU RGB-D 120. This is because the NTU RGB-D 60, and NTU RGB-D 120 datasets are captured in a lab-controlled environment where there are neither illumination changes nor background variations. Conversely, on Toyota-Smarthome,  the RGB video-based network performs similarly or worse than the skeleton-based network due to background clutter and viewpoint variations (see Figure. \ref{fig:motivation}). Also, we can observe that the proposed spatial-temporal attention enhances the accuracy by 0.1\%, 0.2\%, 1.0\%, and 0.9\% for NTU 60 (cross-subject and cross-view evaluation setting) and NTU 120 (XSub and XSet), respectively. In addition, on Toyota-Smarthome dataset, our attention-based multimodal network surpasses its counterpart without such spatial-temporal attention with 1.4\% and 3.0\% in mean class accuracy on CS and CV2 protocols.

\begin{table*}
\caption{Ablations of the proposed multimodal architecture components on three benchmarks: NTU RGB+D 60, NTU RGB+D 120, and Toyota SH. We report the mean Top-1 accuracy(\%) for Toyota-Smarthome dataset and Top-1 accuracy (\%) for other datasets using 1-clip per video. STA stands for spatial-temporal attention}
\label{table:ablation_study_proposed_method}
\centering
\footnotesize
\begin{adjustbox}{width=1\columnwidth,center}
\begin{tabular}{ llllllll} %>{\columncolor[RGB]{167, 199, 231}}l>{\columncolor[RGB]{167, 199, 231}}l
\hline
\multirow{2}{*}{\#}& \multirow{2}{*}{Method} & \multicolumn{2}{c}{NTU 60} &\multicolumn{2}{c}{NTU 120}&\multicolumn{2}{c}{Toyota SH} \\ %& \multicolumn{2}{c}{PKU-MMD}
& & X-Sub & X-View  & X-Sub & X-Set & X-Sub & X-View2\\ \hline  %&X-Sub &X-View

1 & Pose-X-ShiftNet & 92.72 & 96.39& 84.0 &87.26  & 65.74 &59.30 \\ %& 94.27 &97.81
2 & RGB-X-ShiftNet & 94.83& 98.04 & 90.13& 91.58   &62.73 &58.81 \\  %&94.38 &95.87 
 
3& Score fusion (\#1,\#2) & 96.0& 98.77& 91.40 &93.41  &70.29 &64.78 \\%& 94.82 & 98.14

%5& Score fusion (\#1,\#3) & 95.92&97.59&-&-\\
4 & Ours with STA (\#1,\#2) &\textbf{96.1} & \textbf{99.0}& \textbf{92.4}& \textbf{94.3}& \textbf{71.7} & \textbf{67.8} \\ %&\textbf{96.19}& \textbf{98.41} 
% & Ours with STA (RGB-X3D) &96.47 & 98.97& & \\ \hline
\hline
\end{tabular}
\end{adjustbox}
\end{table*}
 
 This confirms that the proposed multimodal method takes full advantage of the complementarity of RGB and skeleton modalities.  

\subsection{Choosing the proportion shift of the X-ShiftNet}
\label{sec:portion_shift}
Since X-ShiftNet utilizes the Temporal Shift Module (TSM) for temporal modeling, we need to study the impact of the different proportion shifts on the X-ShiftNet performance. From Table \ref{table:ablation_study_portion_shift}, we can observe the following: 1) The Pose-X-ShiftNet reaches its peak performance when shifting $1/2$ feature channels between neighboring frames ($1/4$ in each direction). This confirms the intuition that the pose stream focuses on modeling the action dynamics since the included information in the skeleton sequence is the skeleton joint coordinates. 2) X-ShiftNet for RGB stream achieves similar performance when $1/2$ or $1/4$ of feature channels are shifted, although the proportion shift $1/2$ has a higher data movement overhead than that of $1/4$. Therefore, we use proportion shifted channels $1/4$ for the RGB stream and $1/2$ for the pose stream for the rest of the paper.
\begin{table}
\caption{Ablation study of proportion shift's impact on X-ShiftNet performance, evaluated on NTU RGB+D 60's X-Sub and X-View protocols.}
\label{table:ablation_study_portion_shift}
\centering
\footnotesize
\begin{tabular}{ lllll}
\hline
\multirow{2}{*}{Portion shift}& \multicolumn{2}{c}{RGB Stream} & \multicolumn{2}{c}{Pose stream} \\
&  X-Sub & X-View  & X-Sub & X-View  \\ \hline 
$1/4$ & 94.83 &\textbf{98.04}&92.6 &95.7  \\
$1/2$ & \textbf{ 95.03}  &97.98  & \textbf{92.7}  &\textbf{96.4}\\ \hline
\end{tabular}
\end{table}

 \subsubsection{Choosing the appearance and pose network architectures}
A comparison of RGB video-based methods is shown in Table \ref{table:ablation_study_RGB_methods}. We can observe that the proposed RGB-X-ShiftNet outperforms TSM  \cite{lin2019tsm} and the X3D network \cite{feichtenhofer2020x3d} and obtains similar results compared to the SlowOnly network \cite{Duan_poseconv3d}, while requiring 14.5x, 1.1x, and 9.3x fewer GFLOPs. %achieves comparative top-1 accuracy with  8.5-13.3x reduction in FLOPs in comparison with SlowOnly \cite{Duan_poseconv3d} and TSM \cite{lin2019tsm}. 
 For the pose network, it is noticeable from Table \ref{table:ablation_study_pose_methods} that the proposed Pose-X-ShiftNet network performs better than C3D \cite{Duan_poseconv3d} while requiring 4.7x fewer FLOPs. Also, it achieves competitive performance with the X3D-s network\cite{feichtenhofer2020x3d}, and SlowOnly \cite{Duan_poseconv3d} while requiring 1.1x and 4.4x less GFLOPs.

 \begin{table}
\caption{Comparison with RGB video-based methods on NTU RGB+D 60 X-Sub Protocol.}
\label{table:ablation_study_RGB_methods}
\footnotesize
\centering
\begin{tabular}{ lllll}
\hline
Backbone&$T_{RGB}$ &NTU60-XSub& FLOPs& Params\\ \hline
TSM \cite{lin2019tsm} & 8 &92.3 & 33.0G & 23.6M\\
TSM \cite{lin2019tsm}& 16 &93.7 & 65.9G& 23.6M\\
SlowOnly \cite{Duan_poseconv3d} & 8 & 94.9& 42.0G &31.9M \\
X3D \cite{feichtenhofer2020x3d} & 16 &94.0&5.0G& 3.1M\\
\textbf{RGB-X-ShiftNet} (our) & \textbf{16} &\textbf{94.8}&\textbf{4.5G}&\textbf{2.0M}\\ \hline

\end{tabular}
\end{table}
 
\begin{table}
\caption{Comparison with skeleton-based methods on NTU RGB+D 60 X-Sub Protocol.}
\label{table:ablation_study_pose_methods}
\centering
\footnotesize
\begin{tabular}{ lllll}
\hline
Backbone&$T_{pose}$ &NTU60-XSub& FLOPs& Params\\ \hline
C3D \cite{Duan_poseconv3d} & 48 &92.5 & 16.8G & 3.4M\\
SlowOnly \cite{Duan_poseconv3d} & 48 &93.1 & 15.9G& 2.0M\\
X3D \cite{feichtenhofer2020x3d} & 48 &92.84&4.0G& 586.2k\\ 
\textbf{Pose-X-ShiftNet} (our)& \textbf{48}&\textbf{92.72}&\textbf{3.6G}& \textbf{514.7k}\\ \hline
\end{tabular}
\end{table}
 \subsubsection{Choosing the spatial-temporal attention block}
To evaluate the effectiveness of the proposed nesting spatial-temporal attention block, we conducted experiments with the nesting spatial-temporal attention block from \cite{li2021nesting}, which consists of both spatial attention and a nested temporal attention module. The spatial attention module involves a 1x3x3 spatial convolution to compress the number of channels $f_s$ to 1, followed by a 1x7x7 spatial convolution. This can be expressed as: 
  \begin{equation} A_S = \phi(g^{1\mathrm{x}7\mathrm{x}7}(\delta(g^{1\mathrm{x}3\mathrm{x}3}(f_s)))),\end{equation} where $\phi$ and $\delta$ are the Sigmoid and RELU activation functions, respectively.

The temporal attention block, which is inspired by the Squeeze and Excitation(SE) block \cite{hu2018squeeze}, aggregates spatial information of the spatial attention map $A_S$ using global average pooling and then models the temporal-wise dependencies by two fully connected layers with non-linear activation functions (RELU and Sigmoid). This process is formulated as:
\begin{equation}A_T = \phi(W_2(\delta (W_1 (GAP(A_S)))), \end{equation} 
The weights of the fully connected layers are represented by $W_1$ and $W_2$.

Our multimodal network with the proposed nesting spatial-temporal attention block achieves 96.14\% and 99.0\% on the X-Sub and X-View evaluation settings of the NTU RGB-D 60 dataset, respectively, compared to  96.20\% and 98.9\% with the spatial-temporal attention one \cite{li2021nesting}. The similar performance of the two spatial-temporal attention blocks demonstrates the versatility of the proposed multimodal architecture across different attention block designs. Notably, our spatiotemporal attention block is significantly more parameter-efficient, requiring only 107 parameters, compared to the 2.87k parameters used by the attention block in \cite{li2021nesting}. This reduction in parameters highlights the efficiency of our approach without sacrificing the performance.

\subsection{Comparison with state-of-the-art methods}
\begin{table}[h!]
\footnotesize
\caption{ Comparison of top-1 accuracy(\%) with state-of-the-art methods on NTU RGB+D 60 and NTU RGB+D 120 Toyota SmartHome. \textsuperscript{\textdagger} indicates the results are obtained using 10 clips per video. $\circ$ means the skeleton is used in training but not in inference. }\label{table:sotr_comparison} %$V_{12}^3$ denotes that views 1 and 2 are used for training, and view 3 for testing.
%\vspace{-5pt}
\centering
\begin{adjustbox}{width=\columnwidth,center}
\begin{tabular}{llllllllll}
\hline
\multirow{2}{*}{Method} & \multicolumn{2}{c}{Modality} & \multicolumn{3}{c}{NTU 60} &\multicolumn{2}{c}{NTU 120} &GFLOPs& Param (M)\\%&\multicolumn{2}{c}{NTU 120}  \\
 & Skeleton &RGB & XSub & XView & Average  & XSub & XSet & & \\ \hline%& XSub & XSet \\ \hline
ST-GCN \cite{yan2018spatial}& \checkmark & - & 81.5 & 88.3 &84.6 & 79.0&81.3&3.8&3.1 \\
2s-AGCN \cite{shi2019two}& \checkmark & - & 88.5 & 95.1 &91.8&82.9 &84.9&8.8&3.5\\
MS-G3D \cite{liu2020disentangling} & \checkmark & - &  91.5 & 96.2 &93.9&86.9 &88.4&16.7&2.8\\ 
PoseConv3D \cite{Duan_poseconv3d} & \checkmark & - &  93.7 & 96.6 &95.2 &86.0 &89.6&15.9&2.0\\     \hline
C3D \cite{tran2015learning} & - & \checkmark & 63.5&70.3 & 66.9& -& -&38.5&78.4\\ 
I3D-Resnet50 \cite{zhu2019action}& - & \checkmark & 93.2 & 97.7 &95.3 &-&- &51.7&33.0\\  \hline 
STAR-Transformer \cite{ahn2023star}  & \checkmark & \checkmark & 92.0&  96.5&94.3&- & -&&\\
TSMF \cite{bruce2021multimodal} & \checkmark & \checkmark & 92.5&  97.4&95.0&87.0 &89.1&85.4&20.8 \\
VPN \cite{das2020vpn}(I3D) & \checkmark & \checkmark & 93.5& 96.2 &94.6 &86.3 &87.8&107.9&24.0\\
VPN++ \cite{das2021vpn++}  & $\circ$ & \checkmark & 91.9& 94.9 & 93.4 &86.7 &89.3&-&-\\
VPN++ + 3D Poses \cite{das2021vpn++}  & \checkmark & \checkmark & 94.9& 98.1 & 96.5 &90.7 &92.5&125.8&15.5\\
TCEM-MMNet \cite{liu2024temporal}& \checkmark & \checkmark &94.3& \underline{98.8} & 96.6&- &- &-&-\\
MMNet \cite{bruce2022mmnet}(Inception-v3) & \checkmark & \checkmark &\underline{95.3}& 98.4 & 96.8&\textbf{92.9}&\textbf{94.4}&89.2&34.2\\
$\pi$-ViT\textsuperscript{\textdagger} \cite{reilly2024just} & $\circ$ & \checkmark &94.0& 97.9 & 96.0&91.9&92.9 &590.0&121.4\\
Our proposed approach & \checkmark & \checkmark & \textbf{96.1}& \textbf{99.0}&\textbf{97.6} &\underline{92.4}&\underline{94.3}&\textbf{8.1}&\textbf{2.5}\\  \hline
\vspace{-30pt}
\end{tabular}
\end{adjustbox}
\end{table}

\begin{table}
\footnotesize
\caption{  Comparison of mean top-1 accuracy (\%) with state-of-the-art methods on Toyota SmartHome. \textsuperscript{\textdagger} indicates the results are obtained using 10 clips per video. $\circ$ means the skeleton is used in training but not in inference. }
\label{tab:sota_comparison_toyota}
    
    \begin{adjustbox}{width=0.7\columnwidth,center}
    \begin{tabular}{llllll>{\columncolor[gray]{.8}[0pt]}l}
    \hline
    \multirow{2}{*}{Method} & \multicolumn{2}{c}{Modality} & \multicolumn{3}{c}{Toyota SmartHome}\\ %&\multicolumn{2}{c}{NTU 120}  \\
 & Skeleton &RGB & CS & CV2 & Average \\ \hline%& XSub & XSet \\ \hline
TSMF \cite{bruce2021multimodal} & \checkmark & \checkmark &53.8&28.9 &41.4\\
VPN \cite{das2020vpn}(I3D) & \checkmark & \checkmark & 65.2&54.1& 59.7 \\
VPN++ \cite{das2021vpn++}  & $\circ$ & \checkmark & 69& 54.9 & 62.0\\
VPN++ + 3D Poses \cite{das2021vpn++}  & \checkmark & \checkmark &71.0& 58.1&64.6 \\
MMNet \cite{bruce2022mmnet}(ResNet18) & \checkmark & \checkmark &65.1&33.4 & 49.3\\
$\pi$-ViT\textsuperscript{\textdagger} \cite{reilly2024just} & $\circ$ & \checkmark &72.9& 64.8& 68.9\\
$\pi$-ViT + 3D Poses\textsuperscript{\textdagger} \cite{reilly2024just} & $\checkmark$ & \checkmark &\textbf{73.1}&\underline{65.0} & 69.1 \\
Our proposed approach & \checkmark & \checkmark &\underline{71.7}&\textbf{67.8} & \textbf{69.8}\\ \hline
    \end{tabular}
 \end{adjustbox}
\end{table}

\begin{table}
 
\footnotesize
\caption{ Comparison of top-1 accuracy(\%) with state-of-the-art methods on the PKU-MMD dataset. }
\label{tab:sota_comparison_pku}
    
    \begin{adjustbox}{width=0.7\columnwidth,center}
    \begin{tabular}{llllll}
    \hline
    \multirow{2}{*}{Method} & \multicolumn{2}{c}{Modality} & \multicolumn{2}{c}{PKU-MMD}\\ %&\multicolumn{2}{c}{NTU 120}  \\
 & Skeleton &RGB & XSub & XView \\ \hline%& XSub & XSet \\ \hline
STA-LSTM \cite{song2018spatio} & \checkmark &  &86.9&92.6 \\
CNN-based \cite{li2017skeleton} & \checkmark &  &90.4&93.7 \\
SRNet \cite{nie2019srnet} & \checkmark &  &93.1&97.0 \\ \hline
TSMF \cite{bruce2021multimodal} & \checkmark & \checkmark &95.8&97.8 \\
MMNet \cite{bruce2022mmnet}(ResNet18) & \checkmark & \checkmark &96.3 &98.0\\
MMNet \cite{bruce2022mmnet}(Inception-v3) & \checkmark & \checkmark &\textbf{97.2}&\underline{98.1}\\
TCEM-MMNet \cite{liu2024temporal}& \checkmark & \checkmark &\underline{96.8} & 98.0 \\
Our proposed approach & \checkmark & \checkmark &96.2&\textbf{98.4}\\ \hline
    \end{tabular}
 \end{adjustbox}
\end{table}

In Table \ref{table:sotr_comparison}, we compare the performance of the proposed multimodal method with previous single-modal and multimodal methods on NTU-RGB+D 60 and NTU RGB+D 120 datasets. For the NTU-RGB+D 60 dataset, it is noticeable that our approach outperforms and competes with previous skeleton-based, RGB video-based, and multimodal methods. In particular, our approach boosts the average accuracies of state-of-the-art skeleton-based, RGB video-based, and multimodal-based by 2.4\%, 2.3\%, and 0.8\%, respectively. It also surpasses $\pi$-ViT, a distillation-based approach, by 1.2\% average accuracy and achieves competitive performance on X-Sub and X-view protocols with 96.14\% and 99.0\% compared to 96.3\% and 99.0\% of $\pi$-ViT + 3D Poses. The proposed multimodal approach outperforms $\pi$-ViT and matches the performance of $\pi$-ViT + 3D Poses, while using 1-clip per video at testing compared to their 10 clips. Furthermore, our method leverages efficient CNN-based backbones for both visual and pose streams, unlike the Transformer-based architectures of $\pi$-ViT and  $\pi$-ViT+ 3D Poses.   

For the NTU RGB-D 120 dataset, we can notice that the proposed multimodal method achieves comparable top-1 accuracy with other methods while drastically decreasing FLOPs and the number of network parameters. Specifically, FLOPs of our method are less than those of TSMF \cite{bruce2022mmnet}, MMNet \cite{bruce2022mmnet}, and $\pi$-ViT \cite{reilly2024just} with 10.5x, 11.0x, and 72.8x, respectively. This shows that the proposed network meets the real-time requirement of a practical HAR system.

For Toyota-Smarthome dataset (Table \ref{tab:sota_comparison_toyota}), the proposed method surpasses prior works on cross-view2 (CV2) evaluation protocol with large margins (3.0\%) and obtains the second-best mean top-1 accuracy on the cross-subject (CS) protocol. In particular, our approach enhanced the average accuracy of state-of-the-art Transformer-based $\pi$-ViT + 3D Poses \cite{reilly2024just} by 0.7\%. The performance improvement may stem from utilizing 2D poses estimated with HRNet \cite{sun2019deep}, which are of better quality than the 3D poses employed by $\pi$-ViT + 3D Poses \cite{reilly2024just}, especially in the spontaneous activity-driven home environment. The lightweight and competitive performance of the proposed multimodal method makes it suitable for the recognition of activities of daily living. 
 
For PKU-MMD \cite{liu2017pku} (Table \ref{tab:sota_comparison_pku}), our EPAM-Net surpasses existing skeleton-based methods and achieves competitive performance against multimodal-based methods. Specifically, our EPAM-Net enhances the results of skeleton-based SRNet \cite{nie2019srnet} by 3.1\% and 1.4\% for X-Sub and X-View evaluation protocols, respectively. Regarding multimodal-based approaches, it achieves the best performance (98.4\% top-1 accuracy) under the X-View protocol and comparable performance under X-Sub protocols. The competitive performance of our EPAM-Net on the PKU-MMD dataset is consistent with that on the NTU RGB-D 60, RGB-D 120, and Toyota-Smarthome datasets, indicating the generalizability of our approach.

\begin{figure}
    \centering
     \begin{subfigure}[b]{0.48\textwidth}
         \centering
         \includegraphics[width=\textwidth]{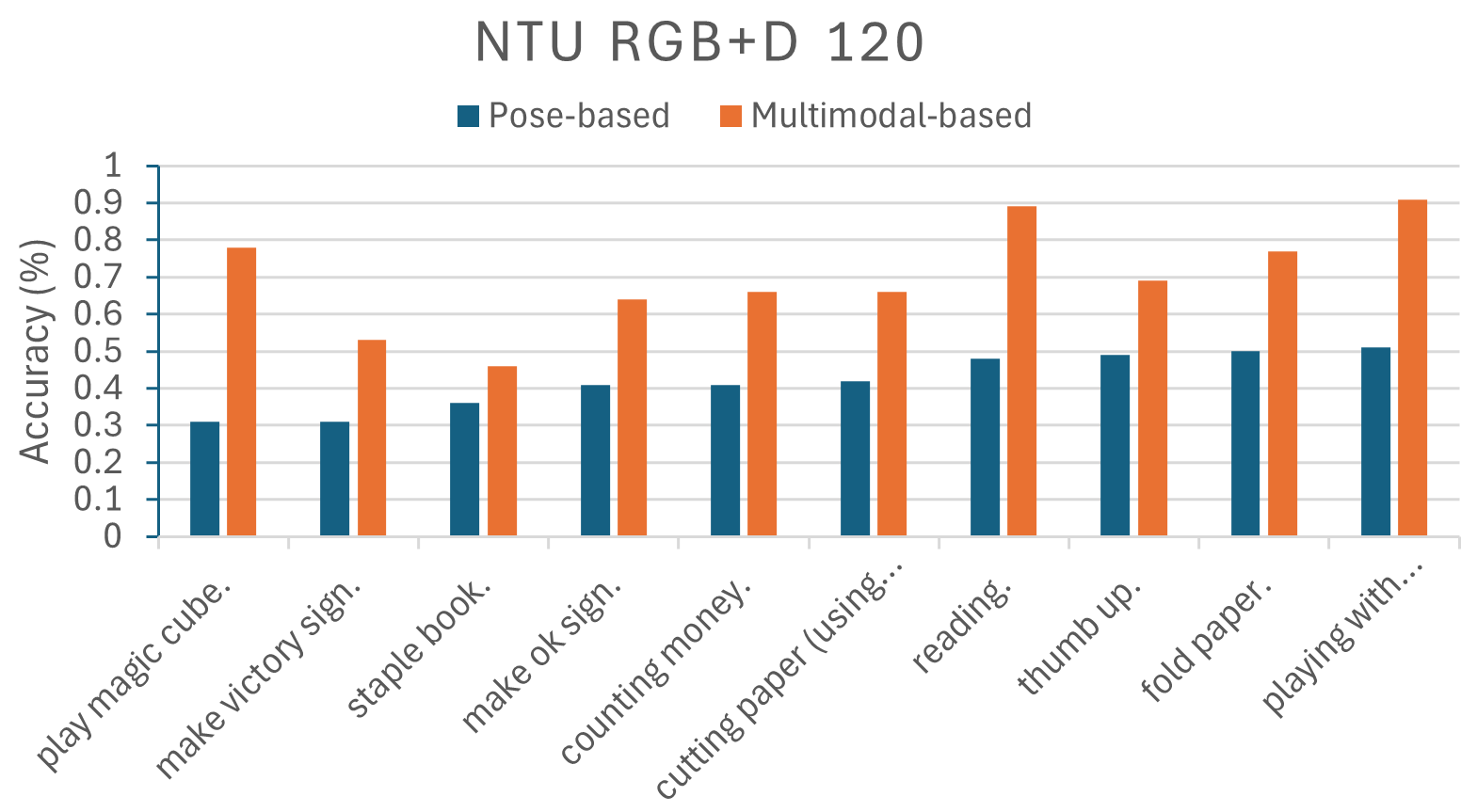}
         \caption{}
         \label{fig:ntu120_confused_actions}
     \end{subfigure}
     \hfill
     \begin{subfigure}[b]{0.48\textwidth}
         \centering
         \includegraphics[width=\textwidth]{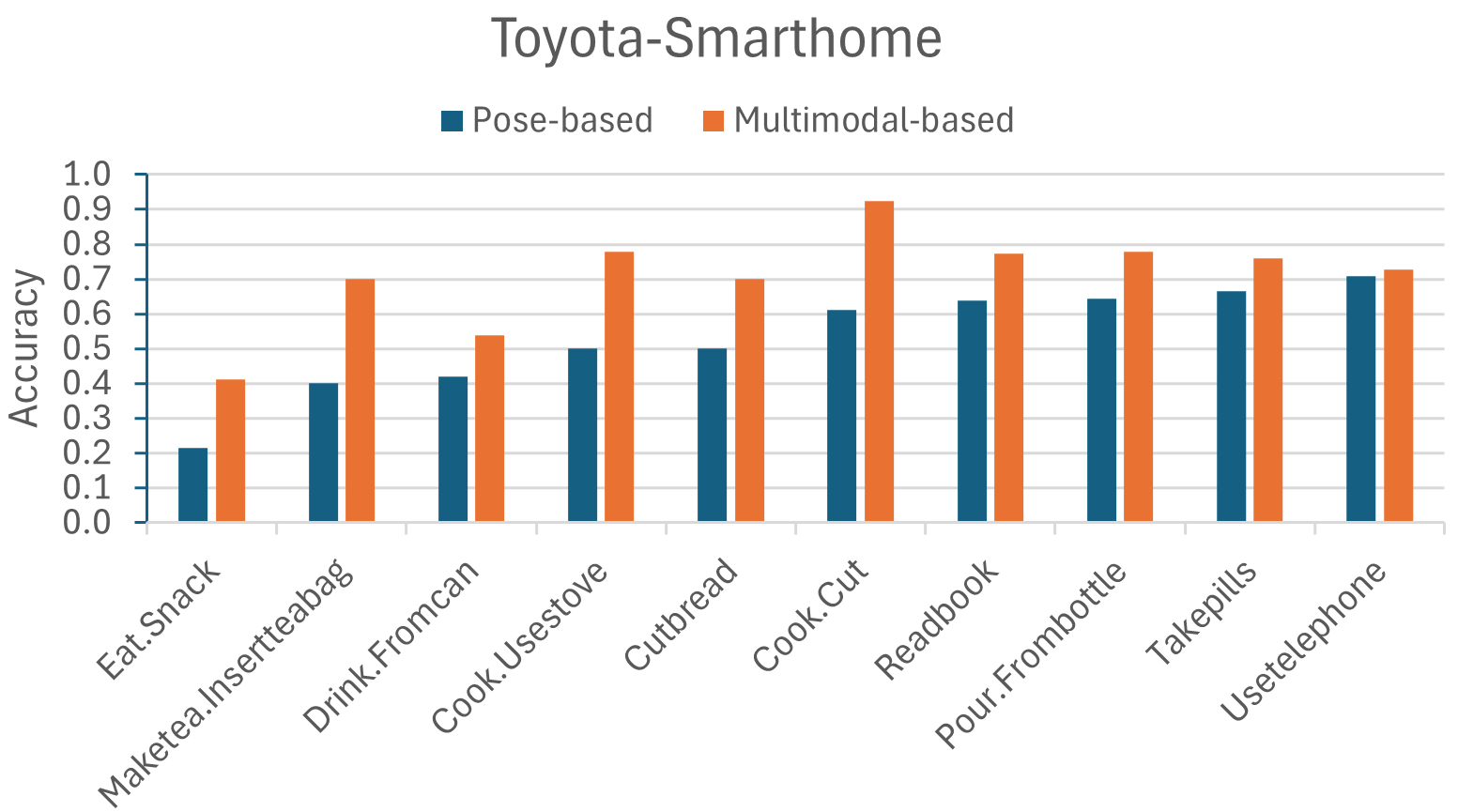}
         \caption{}
         \label{fig:toyta_confused_actions}
     \end{subfigure}
 
    \caption{Classification accuracy per action for top-10 challenging actions on  across-subject protocol of NTU RGB+D 120 (a) and Toyota-Smarthome (b) datasets. }
    \label{fig:confused_actions}
\end{figure}

Figure. \ref{fig:confused_actions} shows the skeleton-based (Pose-X-ShiftNet) and the proposed multimodal-based action recognition accuracy per action for the difficult actions (e.g., actions shown in Figure.\ref{fig:motivation}) on the X-Sub protocol of NTU RGB+D 120 and Toyota-Smarthome datasets. It is noticeable that the proposed multimodal method enhances the accuracy of such actions significantly compared to the pose-X-ShiftNet method.

\section{Conclusion}
Throughout this work, we addressed the extensive computation problem of existing multimodal action recognition networks by introducing a novel and efficient pose-driven attention-guided multimodal network that competes and is computationally efficient with the state of the art. The proposed network utilizes an efficient X-ShiftNet network to model the spatial and temporal information from RGB frames and their corresponding pose sequences. The spatial-temporal attention block guides the visual features to pay attention to the keyframes and their salient spatial regions using the pose features. The final classification score is obtained using the fused predictions of the RGB and the skeleton streams. NTU-60, NTU-120, PKU-MMD, and Toyota-Smarthome datasets showed competitive performance of the proposed method compared to state-of-the-art ones with up to 72.8x reduction in FLOPs and up to 48.6x reduction in the number of network parameters.

 \bibliographystyle{elsarticle-num} 
 %\bibliography{refs}

\end{document}